\documentclass[journal]{IEEEtran}
\usepackage{setspace}
\usepackage[section]{placeins}
\usepackage{graphicx,subfigure}
\usepackage{amsmath}
\usepackage{amsfonts}
\usepackage{caption2}
\usepackage{amssymb}
\usepackage{varwidth}
\usepackage{multirow}
\usepackage{booktabs}
\usepackage[usenames, dvipsnames]{color}
\usepackage[linesnumbered,ruled,vlined]{algorithm2e}
\usepackage{array}
\usepackage{rotating}
\usepackage{epsfig,amsfonts,multirow}
\usepackage{bm}
\usepackage{cite}

\begin{document}
\title{Cascaded Recurrent Neural Networks for Hyperspectral Image Classification}
\author{Renlong~Hang,~\IEEEmembership{Member,~IEEE}, Qingshan~Liu,~\IEEEmembership{Senior~Member,~IEEE}, \\
Danfeng Hong, and Pedram Ghamisi,~\IEEEmembership{Senior~Member,~IEEE}

\thanks{This work was supported in part by the Natural Science Foundation of China under Grants 61532009 and 61825601, in part by the Natural Science Foundation of Jiangsu Province, China, under Grant BK20180786. (Corresponding author: Qingshan Liu.)

R. Hang and Q. Liu are with the Jiangsu Key Laboratory of Big Data Analysis Technology, the School of Automation, Nanjing University of Information Science and Technology, Nanjing 210044, China (renlong\_hang$@$163.com, qsliu$@$nuist.edu.cn).

D. Hong is with the Remote Sensing Technology Institute (IMF), German Aerospace Center (DLR), 82234 Wessling, Germany, and Signal Processing in Earth Observation (SiPEO), Technical University of Munich (TUM), 80333 Munich, Germany(e-mail: danfeng.hong@dlr.de).

P. Ghamisi is with the Helmholtz-Zentrum Dresden-Rossendorf (HZDR), Helmholtz Institute Freiberg for Resource Technology (HIF), Exploration, D-09599 Freiberg, Germany (e-mail: p.ghamisi@gmail.com). }}

\maketitle

\begin{abstract}
\textcolor{blue}{This paper has been accepted by IEEE Transactions on Geoscience and Remote Sensing.} By considering the spectral signature as a sequence, recurrent neural networks (RNNs) have been successfully used to learn discriminative features from hyperspectral images (HSIs) recently. However, most of these models only input the whole spectral bands into RNNs directly, which may not fully explore the specific properties of HSIs. In this paper, we propose a cascaded RNN model using gated recurrent units (GRUs) to explore the redundant and complementary information of HSIs. It mainly consists of two RNN layers. The first RNN layer is used to eliminate redundant information between adjacent spectral bands, while the second RNN layer aims to learn the complementary information from non-adjacent spectral bands. To improve the discriminative ability of the learned features, we design two strategies for the proposed model. Besides, considering the rich spatial information contained in HSIs, we further extend the proposed model to its spectral-spatial counterpart by incorporating some convolutional layers. To test the effectiveness of our proposed models, we conduct experiments on two widely used HSIs. The experimental results show that our proposed models can achieve better results than the compared models.

\end{abstract}
\begin{IEEEkeywords}
Recurrent neural network (RNN), gated recurrent unit (GRU), spectral feature, spectral-spatial feature, hyperspectral image (HSI) classification.

\end{IEEEkeywords}
\IEEEpeerreviewmaketitle

\section{Introduction}
With the development of imaging technology, current hyperspectral sensors can fully portray the surface of the Earth using hundreds of continuous and narrow spectral bands, ranging from the visible spectrum to the short-wave infrared spectrum. The generated hyperspectral image (HSI) is often considered as a three-dimensional cube. The first two are spatial dimensions, which record the locations of each object. The third one is spectral dimension, which captures the spectral signature (reflective or emissive properties) of each material in different bands along the electromagnetic spectrum \cite{ghamisi2017}. Using such rich information, HSIs have been widely applied to various applications, such as land cover/land use classification, precision agriculture, and change detection. For these applications, one basic but important procedure is HSI classification, whose goal is to assign candidate class labels to each pixel.

In order to acquire accurate classification results, numerous methods have been proposed. For example, one can directly consider the rich spectral signature as features and feed them into advanced classifiers, such as support vector machine (SVM) \cite{mountrakis2011}, random forest \cite{belgiu2016} and extreme learning machine \cite{li2015}. However, due to the dense spectral sampling of HSIs, there may exist some redundant information among adjacent spectral bands. This easily leads to the so-called curse of dimensionality (the Hughes effect) which causes a sudden drop in classification accuracy when there is no balance between the high number of spectral channels and a limited number of training samples. Therefore, a large number of works were proposed to mine discriminative features from the high-dimensional spectral signature \cite{jia2013}. Popular models include principle component analysis (PCA), linear discriminant analysis (LDA) \cite{liao2013,hang2016,hang2017}, and graph embedding \cite{lunga2014,hang2018,zhao2018land}. Besides, representation-based models have also been employed to HSI classification in recent years. In \cite{chen2011} and \cite{fang2014}, sparse representation was proposed to learn discriminative features from HSIs. Similarly, collaborative representation was also widely explored \cite{li2014},\cite{liw2014}. In these models, an input spectral signature is usually represented by a linear combination of atoms from a dictionary, and the classification result can be derived from the reconstructed residual without needing to train extra classifiers, which often costs much time.

Although the aforementioned models have demonstrated their effectiveness in the field of HSI classification, there still exist some drawbacks to address. For the traditional feature extraction models, we need to pre-define a mining criterion (e.g., maximizing the between-class scatter matrix in LDA), which heavily depends on the domain knowledge and experience of experts. For the representation-based models, their goal is to reconstruct the input signal, leading to sub-optimal representation for classification. Additionally, all of them can be considered as shallow-layer models, which limit their potentials to learn high-level semantic features. Recently, deep learning \cite{lecun2015},\cite{schmidhuber2015}, a very hot research topic in machine learning, has shown its huge superiority in most fields of computer vision \cite{krizhevsky2012,he2016,girshick2014,ren2015} and natural language processing \cite{collobert2011},\cite{sutskever2014}. The goal of deep learning is to learn nonlinear, high-level semantic features from data in a hierarchical manner.

Due to the effects of multi-path scattering and the heterogeneity of sub-pixel constituents, HSI often lies in a nonlinear and complex feature space. Deep learning can be naturally adopted to deal with this issue \cite{zhang2016},\cite{zhu2017}. In the past few years, many deep learning models were successfully applied to HSI classification. For example, in \cite{chen2014,tao2015,ma2016}, the autoencoder model has been used to learn deep features from high-dimensional spectral signature directly. Similar to autoencoder, deep belief network was also explored to extract spectral features \cite{chen2015,zhou2017,zhong2017}. However, both of them belong to fully-connected networks, which contain large numbers of parameters to train. Different from them, convolutional neural networks (CNNs) have local connection and weight sharing properties, thus largely reducing the number of training parameters \cite{li2017hyperspectral,zhao2017object,zhang2018}. In \cite{hu2015}, Hu \textit{et al}. proposed to use one-dimensional CNN to learn and represent the spectral information. This model is comprised of an input layer, a convolutional layer, a pooling layer, a fully-connected layer and an output layer. The whole model is trained in an end-to-end manner, thus achieving satisfying results for HSI classification.

Besides spectral information, HSIs also have rich spatial information. How to combine them together has been an active research topic in the field of HSI classification \cite{he2018},\cite{ghamisi2018}. One potential method is to extend the spectral classification model into its spectral-spatial counterpart. For instance, in \cite{chen2016,li2017,shi2018}, a three-dimensional CNN was employed to spectral-spatial classification of HSIs. However, due to the simultaneous convolution operators in both spectral domain and spatial domain, the computational complexity is dramatically increased.
In addition, the number of trainable parameters in three-dimensional CNNs is also a problem.
In order to perform three-dimensional convolution, the dimensionality of the input and the dimensionality of the kernel (filter) should be equal. This heavily increases the number of parameters. Another candidate method for spectral-spatial classification is the one based on two-branch networks. One branch is for spectral classification and the other one for spatial classification. In \cite{yang2017,xu2018,hao2018}, one-dimensional CNN or autoencoder was used to learn spectral features and two-dimensional CNN was designed to learn spatial features. These two features are then integrated together via feature-level fusion or decision-level fusion. For two-dimensional CNNs, only a few principal components were extracted and used as inputs, thus reducing the computational consuming compared to three-dimensional CNNs.

Most of existing models can be considered as vector-based methodologies. Recently, a few works attempted to regard HSIs as sequential data, so recurrent neural networks (RNNs) were naturally used to learn features. In \cite{wu2017}, Wu \textit{et.al} proposed using RNN to extract spectral features from HSIs. In \cite{liu2017} and \cite{zhou2018}, a variant of RNN using long short-term memory (LSTM) units was designed to learn spectral-spatial features from HSIs. In \cite{zhou2018Integrating}, another variant of RNN using gated recurrent units (GRUs) was employed. Compared to the widely explored CNN models, RNNs have many superiorities. For example, the key component of CNNs is the convolutional operator. Due to the kernel size limitations of it, one-dimensional CNNs can only learn the local spectral dependency while easily ignoring the effects of non-adjacent spectral bands. Different from them, RNNs, especially using GRU or LSTM, often input spectral bands one by one via recurrent operators, thus capturing the relationship from the whole spectral bands. Besides, RNNs often have smaller numbers of parameters to train than CNNs, so they will be more efficient in the training and inferring phases.

Benefiting from its powerful learning ability from sequential data, current RNN-related models often simply input the whole spectral bands to networks, which may not fully explore the redundant and complementary properties of HSIs. The redundant information between adjacent spectral bands will increase the computational burden of RNNs without improving the classification results. Sometimes such redundancy may reduce the classification accuracy since it increases within-class variances and decreases between-class variances in the feature space. Besides, it may also increase the difficulties in learning complementary information. To address these issues, we propose a cascaded RNN model using gated recurrent units (GRUs) in this paper. This model mainly consists of two RNN layers. The first RNN layer focuses on reducing the redundant information of adjacent spectral bands. These reduced information are then fed into the second RNN layer to learn their complementary features. Besides, in order to improve the discriminative ability of the learned features, we design two strategies for the proposed model. Finally, we also extend the proposed model to its spectral-spatial version by incorporating some convolutional layers. The major contributions of this paper are summarized as follows.
\begin{enumerate}
  \item We propose a cascaded RNN model with GRUs for HSI classification. Compared to the existing RNN-related models, our model can sufficiently consider the redundant and complementary information of HSIs via two RNN layers. The first one is to reduce redundancy and the second one is to learn complementarity. These two layers are integrated together to generate an end-to-end trainable model.
  \item In order to learn more discriminative features, we design two strategies to construct connections between the first RNN layer and the output layer. The first strategy is the weighted fusion of features from two layers, and the second one is the weighted combination of different loss functions from two layers. Their weights can be adaptively learned from data itself.
  \item To capture the spectral and spatial features simultaneously, we further extend the proposed model to its spectral-spatial counterpart. A few convolutional layers are integrated into the proposed model to learn spatial features from each band, and these features are then combined together via recurrent operators.
\end{enumerate}

The rest of this paper is organized as follows. Section II describes the details of the proposed models, including a brief introduction of RNN, and the structure of the proposed model as well as its modifications. The descriptions of data sets and experimental results are given in Section III. Finally, Section IV concludes this paper.

\section{Methodology}
\begin{figure*}[!t]
  \centering
  \includegraphics[scale = 0.6]{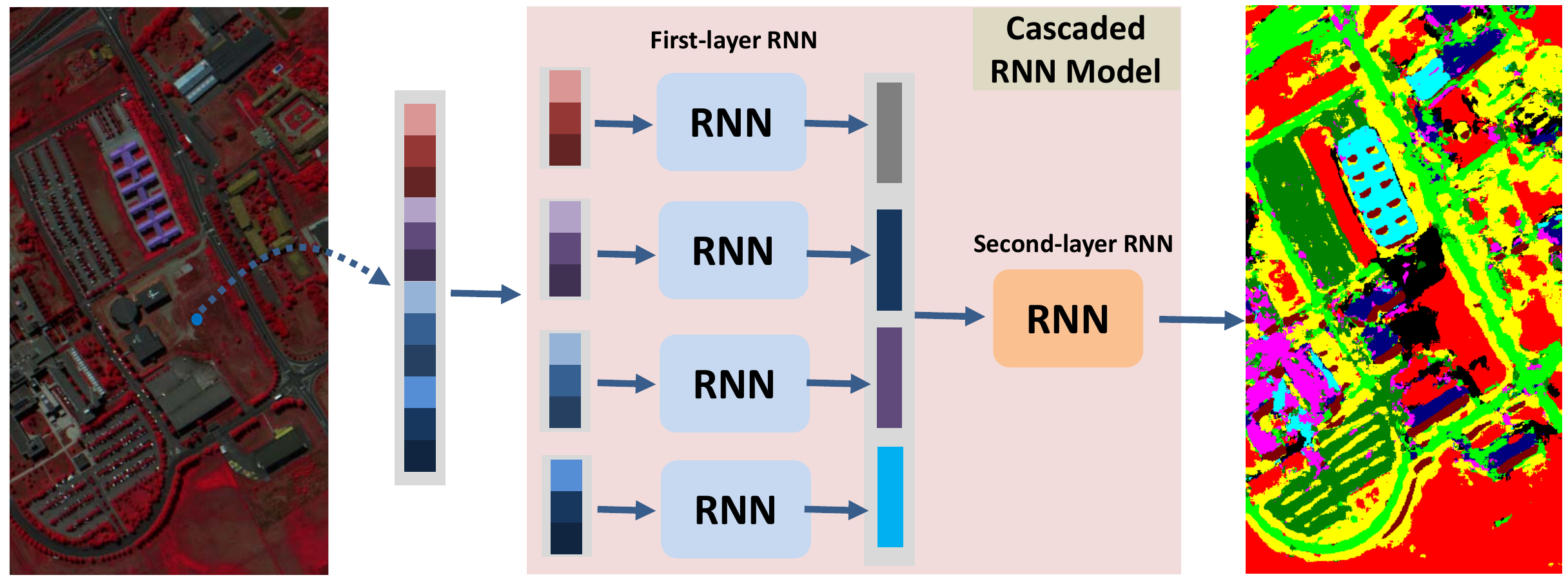}\\
  \caption{Flowchart of the proposed model.}\label{RNNs}
\end{figure*}
As shown in Fig.$~$\ref{RNNs}, the proposed cascaded RNN model mainly consists of four steps. For a given pixel, we firstly divide it into different spectral groups. Then, for each group, we consider the spectral bands in it as a sequence, which is fed into a RNN layer to learn features. After that, the learned features from each group are again regraded as a sequence and fed into another RNN layer to learn their complementary information. Finally, the output of the second RNN layer is connected to a softmax layer to derive the classification result.
\subsection{Review of RNN}
RNN has been widely used for sequential data analysis, such as speech recognition and machine translation \cite{sutskever2014},\cite{graves2013}. Assume that we have a sequence data $\mathbf{x} = (\mathbf{x}_{1}, \mathbf{x}_{2}, \cdots, \mathbf{x}_{T})$, where $\mathbf{x}_{t}, t\in\{1,2,\cdots,T\}$ generally represents the information at the $t$-th time step. When applying RNN to HSI classification, $\mathbf{x}_{t}$ will correspond to the spectral value at the $t$-th band. For RNN, the output of hidden layer at time $t$ is
\begin{equation}\label{hidden}
 \mathbf{h}_{t} = \phi(\mathbf{W}_{hi}\mathbf{x}_{t} + \mathbf{W}_{hh}\mathbf{h}_{t-1} + \mathbf{b}_{h})
\end{equation}
where $\phi$ is a nonlinear activation function such as logistic sigmoid or hyperbolic tangent functions, $\mathbf{b}_{h}$ is a bias vector, $\mathbf{h}_{t-1}$ is the output of hidden layer at the previous time, $\mathbf{W}_{hi}$ and $\mathbf{W}_{hh}$ denote weight matrices from the current input layer to hidden layer and the previous hidden layer to current hidden layer, respectively. From this equation, we can observe that via a recurrent connection, the contextual relationships in the time domain can be constructed. Ideally, $\mathbf{h}_{T}$ can capture most of the time information for the sequence data.

For classification tasks, $\mathbf{h}_{T}$ is often fed into an output layer, and the probability that the sequence belongs to $i$-th class can be derived by using a softmax function. These processes can be formulated as
\begin{equation}
\begin{aligned}
& \qquad \mathbf{O}_{T} = \mathbf{W}_{oh}\mathbf{h}_{T} + \mathbf{b}_{o}\\
& P(\tilde{y}=i|\boldsymbol{\theta},\mathbf{b}) = \frac{e^{\boldsymbol{\theta}_{i}\mathbf{O}_{T}+b_{i}}}{\sum_{j=1}^{C}e^{\boldsymbol{\theta}_{j}\mathbf{O}_{T}+b_{j}}}
\end{aligned}
\end{equation}
where $\mathbf{b}_{o}$ is a bias vector, $\mathbf{W}_{oh}$ is the weight matrix from hidden layer to output layer, $\boldsymbol{\theta}$ and $\mathbf{b}$ are parameters of softmax function, $C$ is the number of classes to discriminate. All of these weight parameters in Equation (1) and (2) can be trained using the following loss function
\begin{equation}\label{Loss}
\mathcal{L}=-\frac{1}{N}\sum_{i=1}^{N}[y_{i}\textmd{log}(\tilde{y}_{i})+
(1-y_{i})\textmd{log}(1-\tilde{y}_{i})]
\end{equation}
where $N$ is the number of training samples, $y_{i}$ and $\tilde{y}_{i}$ are the true label and the predicted label of the $i$-th training sample, respectively. This function can be optimized using a backpropagation through time (BPTT) algorithm.

\subsection{Cascaded RNNs}\label{CasRNNs}
HSIs can be described as a three-dimensional matrix $\mathbf{X}\in \mathfrak{R}^{m\times n\times k}$, where $m$, $n$ and $k$ represent the width, height and number of spectral bands, respectively. For a given pixel $\mathbf{x}\in \mathfrak{R}^{k}$, we can consider it as a sequence whose length is $k$, so RNN can be naturally employed to learn spectral features. However, HSIs often contain hundreds of bands, making $\mathbf{x}$ a very long sequence. Such long-term sequence increases the training difficulty since the gradients tend to either vanish or explode \cite{chung2014}. To address this issue, one popularly used method is to design a more sophisticated activation function by using gating units such as the LSTM unit and GRU \cite{cho2014}. Compared to LSTM unit, GRU has a fewer number of parameters \cite{chung2014}, which may be more suitable for HSI classification because it usually has a limited number of training samples. Therefore, we select GRU as the basic unit of RNN in this paper.

The core components of GRU are two gating units that control the flow of information inside the unit. Instead of using Equation$~$(\ref{hidden}), the activation of the hidden layer for band $t$ is now formulated as
\begin{equation}
\mathbf{h}_{t} = (1-u_{t})\mathbf{h}_{t-1} + u_{t}\tilde{\mathbf{h}}_{t}
\end{equation}
where $u_{t}$ is the update gate, which can be derived by
\begin{equation}
u_{t} = \sigma(w_{u}x_{t} + \mathbf{v}_{u}\mathbf{h}_{t-1})
\end{equation}
where $\sigma$ is a sigmoid function, $w_{u}$ is a weight value, and $\mathbf{v}_{u}$ is a weight vector. Similarly, $\tilde{\mathbf{h}}_{t}$ can be computed by
\begin{equation}
  \tilde{\mathbf{h}}_{t} = tanh(\mathbf{w}x_{t} + \mathbf{V}(\mathbf{r}_{t}\odot \mathbf{h}_{t-1}))
\end{equation}
where $\odot$ denotes an element-wise multiplication, and $\mathbf{r}_{t}$ is the reset gate, which can be derived by
\begin{equation}
\mathbf{r}_{t} = \sigma(\mathbf{w}_{r}x_{t} + \mathbf{V}_{r}\mathbf{h}_{t-1})
\end{equation}

Due to the dense spectral sampling of hyperspectral sensors, adjacent bands in HSIs have some redundancy while non-adjacent bands have some complementarity. In order to take account of such information comprehensively, we propose a cascaded RNN model. Specifically, we divide the spectral sequence $\mathbf{x}$ into $l$ sub-sequences $\mathbf{z} = (\mathbf{z}_{1}, \mathbf{z}_{2}, \cdots, \mathbf{z}_{l})$, each of which consists of adjacent spectral bands. Besides the last sub-sequence $\mathbf{z}_{l}$, the length of the other sub-sequences is $d = \textmd{floor}(k/l)$, which denotes the nearest integers less than or equal to $k/l$. Thus, for the $i$-th sub-sequence $\mathbf{z}_{i}, i\in\{1,2,\cdots,l\}$, it is comprised of the following bands
\begin{align}
    \mathbf{z}_{i} = \left\{
   \begin{array}{ll}
    (x_{(i-1)\times d+1}, \cdots, x_{i\times d}), & \hbox{if}\: \;i\neq l,\\
    (x_{(i-1)\times d+1}, \cdots, x_{k}), & \hbox{otherwise}.
   \end{array}
  \right.
\end{align}

\begin{figure}
  \centering
  \includegraphics[scale=0.6]{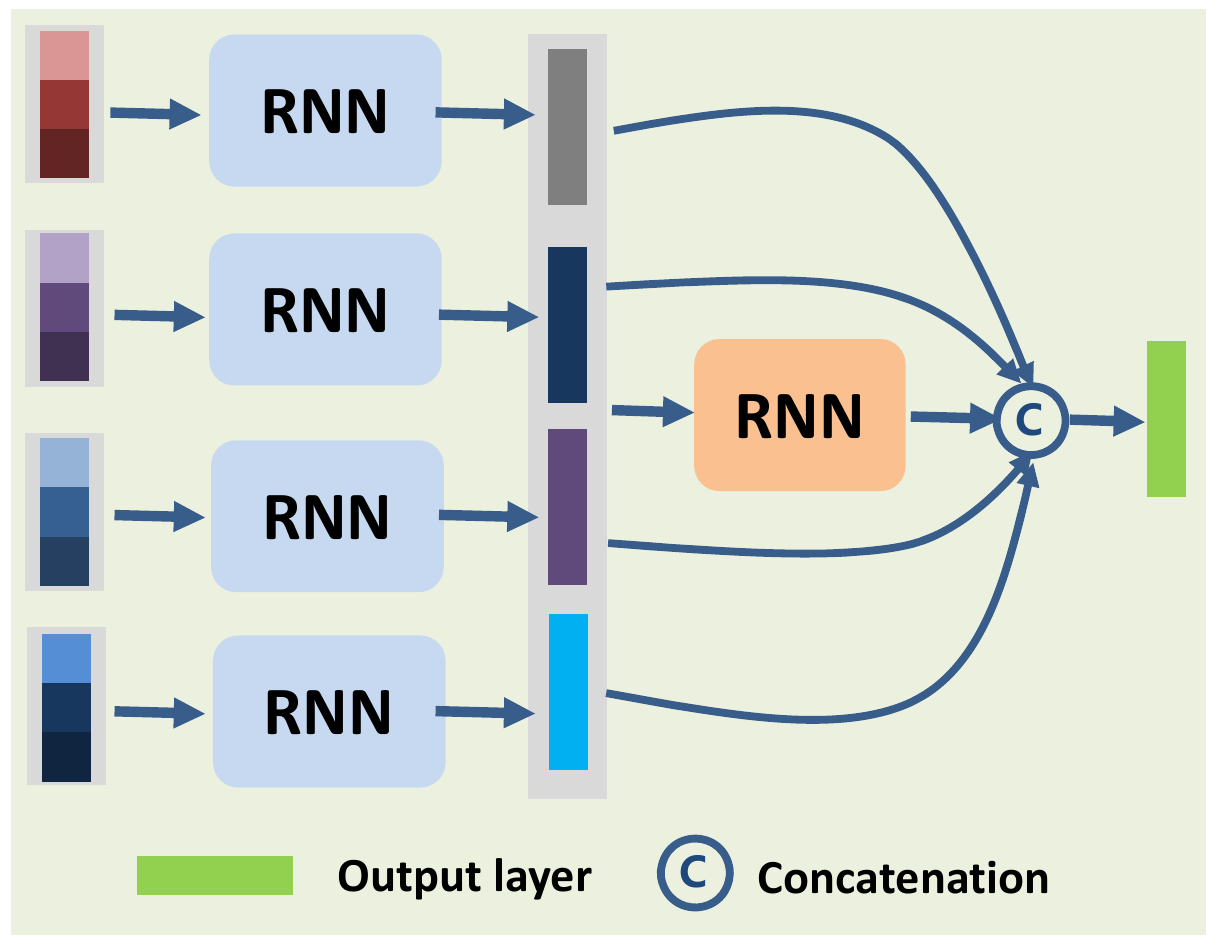}\\
  \caption{The first improvement strategy.}\label{Improve1}
\end{figure}
Then, we feed all the sub-sequences into the first-layer RNNs respectively. These RNNs have the same structure and share parameters, thus reducing the number of parameters to train. In the sub-sequence $\mathbf{z}_{i}$, each band has an output from GRU. We use the output of the last band as the final feature representation for $\mathbf{z}_{i}$, which can be denoted as $\mathbf{F}_{i}^{(1)}\in \mathfrak{R}^{H_{1}}$, where $H_{1}$ is the size of the hidden layer in the first-layer RNN. After that, we can combine $\mathbf{F}_{i}^{(1)}, i\in\{1,2,\cdots,l\}$ together to generate another sequence $\mathbf{F} =(\mathbf{F}_{1}^{(1)}, \mathbf{F}_{2}^{(1)}, \cdots, \mathbf{F}_{l}^{(1)})$ whose length is $l$. This sequence is fed into the second-layer RNN to learn their complementary information. Similar to the first-layer RNNs, we also use the output of GRU at the last time $l$ as the learned feature $\mathbf{F}^{(2)}$. To get a classification result of $\mathbf{x}$, we need to input $\mathbf{F}^{(2)}$ into an output layer whose size equals to the number of candidate classes $C$. Both of these two-layer RNNs have many weight parameters. We choose Equation$~$(\ref{Loss}) as a loss function and use the BPTT algorithm to optimize them simultaneously.

\subsection{Improvement for Cascaded RNNs}

\begin{figure}
  \centering
  \includegraphics[scale=0.6]{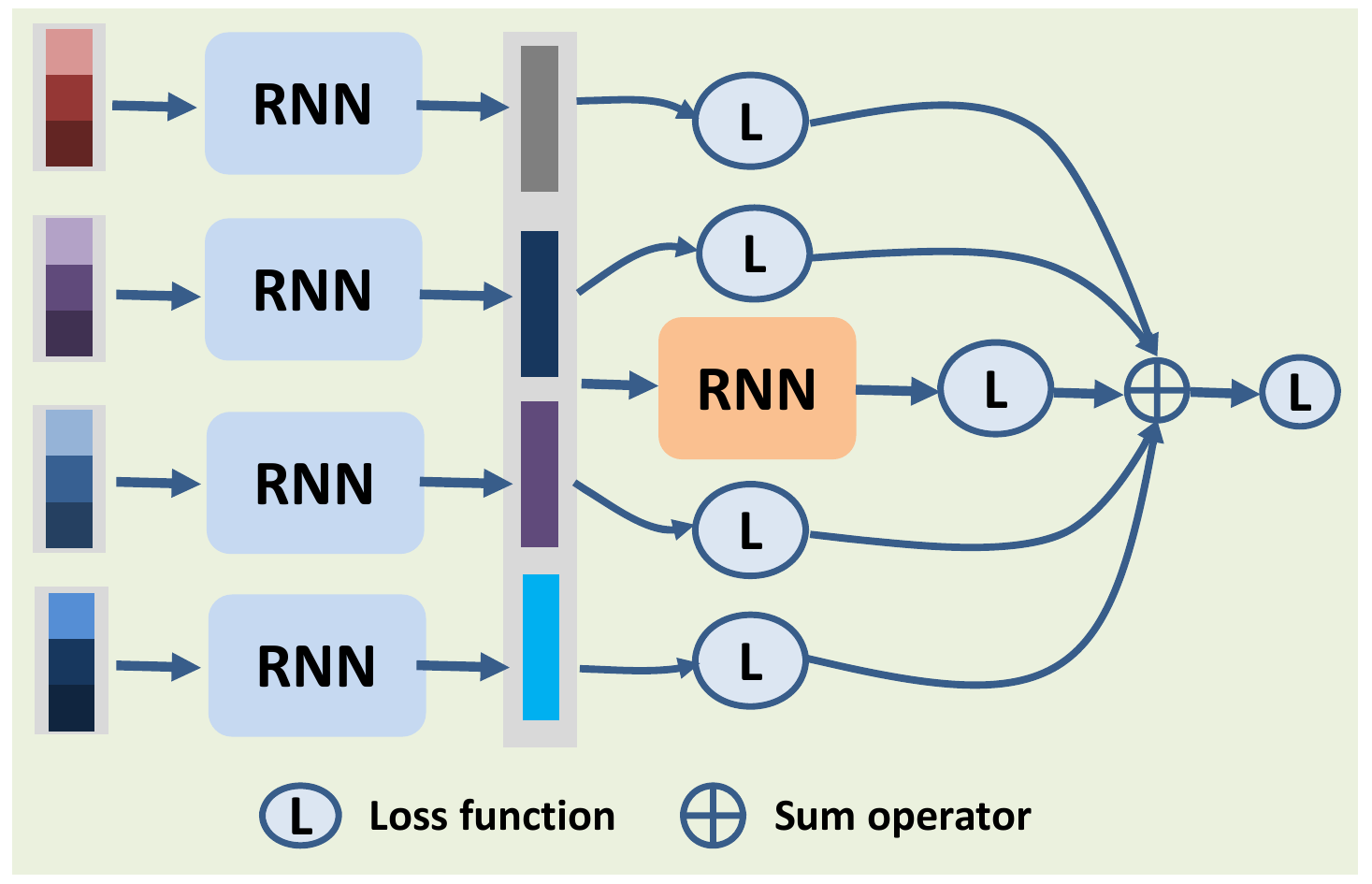}\\
  \caption{The second improvement strategy.}\label{Improve2}
\end{figure}

As described in subsection \ref{CasRNNs}, the second-layer RNN is directly connected to the output layer, so it may be optimized better than the first-layer RNNs. However, the performance of the first-layer RNNs will have effects on the second-layer RNN. In order to improve the discriminative ability of $\mathbf{F}^{(2)}$, an intuitive method is to construct  relations between the first-layer RNNs and the output layer. Here, we propose two strategies to achieve this goal. The first strategy is based on the feature-level connection shown in Fig.$~$\ref{Improve1}. Instead of feeding the output of the second-layer RNN into the output layer only, we attempt to feed all the output features from the first- and the second-layer RNNs in a weighted concatenation manner. Specifically, the input of the output layer is computed as follows
\begin{equation}\label{feature-level}
 \tilde{\mathbf{F}} = [w_{1}^{(1)}\mathbf{F}_{1}^{(1)}, w_{2}^{(1)}\mathbf{F}_{2}^{(1)},\cdots, w_{l}^{(1)}\mathbf{F}_{l}^{(1)}, w^{(2)}\mathbf{F}^{(2)}]
\end{equation}
where $w_{i}^{(1)}\in\mathfrak{R}^{1}, i\in\{1,2,\cdots,l\}$ are fusion weights for the first-layer RNNs, and $w^{(2)}\in\mathfrak{R}^{1}$ is the fusion weight for the second-layer RNN. These weights can be integrated into the whole network and their optimal values are automatically learned from data. The same as the original two-layer RNN model, we also use Equation$~$(\ref{Loss}) to construct the loss function and use the BPTT algorithm to optimize it.

\begin{figure*}
  \centering
  \includegraphics[scale=0.6]{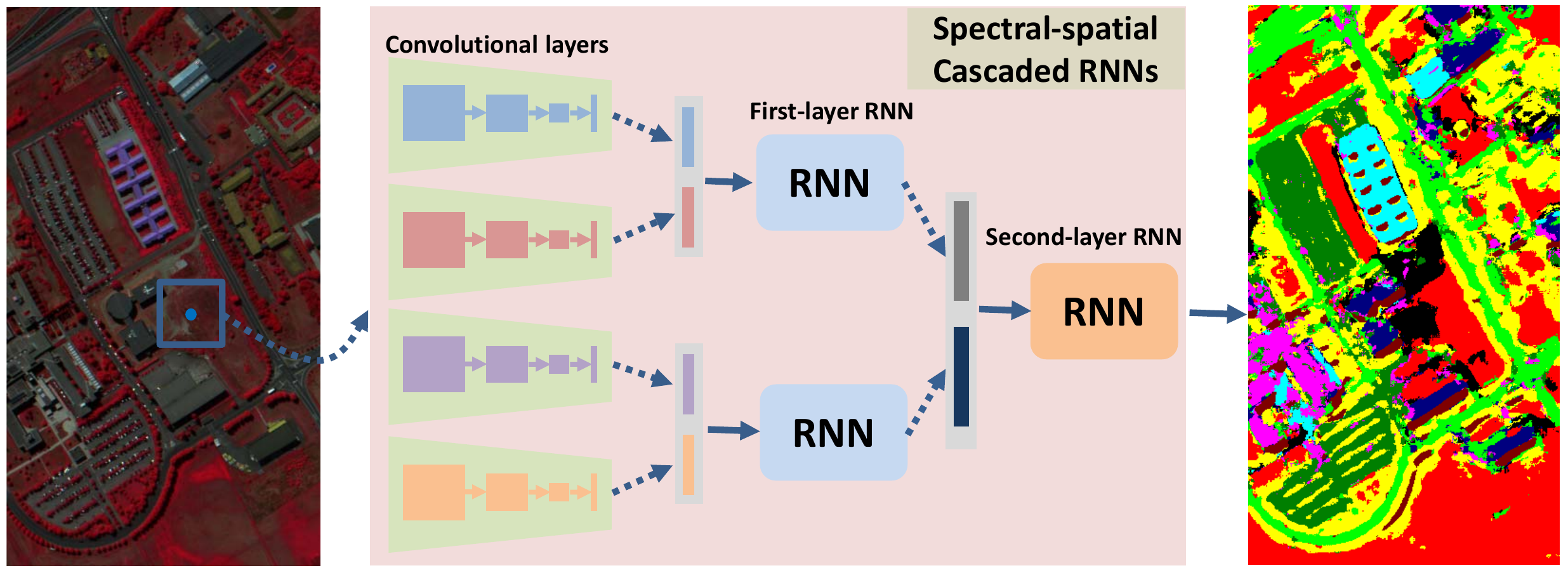}\\
  \caption{Flowchart of spectral-spatial cascaded RNN model.}\label{SSRNN}
\end{figure*}

Different from the first improvement strategy, our second strategy is based on the output-level connection. As shown in Fig.$~$\ref{Improve2}, we feed the features extracted by the first-layer RNNs into output layers, respectively, so that they can learn more discriminative features. Combining these features together using the second-layer RNN will result in a better $\mathbf{F}^{(2)}$. In particular, for $\mathbf{F}_{i}^{(1)}, i\in\{1,2,\cdots,l\}$, we can input it into an output layer and construct a loss function $L_{i}^{(1)}, i\in\{1,2,\cdots,l\}$. Meanwhile, we also input $\mathbf{F}^{(2)}$ into an output layer and construct another loss function $L^{(2)}$. After that, a weighted summation method can be used to combine them together, which can be formulated as
\begin{equation}\label{output-level}
 \tilde{L} = \frac{1}{l}\sum_{i=1}^{l}w_{i}^{(1)}L_{i}^{(1)} + w^{(2)}L^{(2)}
\end{equation}
where $w_{i}^{(1)}\in\mathfrak{R}^{1}$ and $w^{(2)}\in\mathfrak{R}^{1}$ are fusion weights, $L_{i}^{(1)}$ and $L^{(2)}$ are derived from Equation$~$(\ref{Loss}). The final loss function $\tilde{L}$ can be optimized by using the BPTT algorithm. In the prediction phase, we can delete the output layers of the first-layer RNNs and use the output from the second-layer RNN as the final classification result.

\subsection{Spectral-spatial Cascaded RNNs}
Due to the effects of atmosphere, instrument noises, and natural spectrum variations, materials from the same class may have very different spectral responses, while those from different classes may have similar spectral responses. If we only use the spectral information, the resulting classification maps will have many outliers, which is known as the ``salt and pepper'' phenomenon. As a three-dimensional cube, HSIs also have rich spatial information, which can be used as a complement to address this issue. Among numerous deep learning models, CNNs have demonstrated their superiority in spatial feature extraction. In \cite{chen2016}, a typical two-dimensional CNN is designed to extract spatial features from HSIs. The input of this model is the first principle component of HSIs.

Inspired from the two-dimensional CNN model, we extend the cascaded RNN model to its spectral-spatial version by adding some convolutional layers. Fig.$~$\ref{SSRNN} shows the flowchart of the proposed spectral-spatial cascaded RNN model. For a given pixel $\mathbf{x}\in \mathfrak{R}^{k}$, we select a small cube $\hat{\mathbf{x}}\in \mathfrak{R}^{\omega\times\omega\times k}$ centered at it. Then, we split this cube into $k$ matrices $\hat{\mathbf{x}}_{i}\in \mathfrak{R}^{\omega\times\omega}, i\in\{1,2,\cdots, k\}$ across the spectral domain. For each $\hat{\mathbf{x}}_{i}$, we feed it into several convolutional layers to learn spatial features. The same as \cite{chen2016}, we also use three convolutional layers, and the first two layers are followed by pooling layers. The input size $\omega\times\omega$ is $27\times 27$. The sizes of the three convolutional filters are $4\times 4\times 32$, $5\times 5\times 64$ and $4\times 4\times 128$, respectively. After these convolutional operators, each $\hat{\mathbf{x}}_{i}$ will generate a $128$-dimensional spatial feature $\mathbf{s}_{i}$. Similar to the cascaded RNN model, we can also consider $\mathbf{s}=(\mathbf{s}_{1},\mathbf{s}_{2},\cdots,\mathbf{s}_{k})$ as a sequence whose length is $k$. This sequence is divided into $l$ sub-sequences, and they are subsequently fed into the first-layer RNNs respectively to reduce redundancy inside each sub-sequence. The outputs from the first-layer RNNs are combined again to generate another sequence, which are fed into the second-layer RNN to learn complementary information.

Compared to the cascaded RNN model, the spectral-spatial cascaded RNN model is deeper and more difficult to train. Therefore, we propose a transfer learning method to train it. Specifically, we firstly pre-train the convolutional layers using all of $\hat{\mathbf{x}}_{i}, i\in\{1,2,\cdots, k\}$. We replace two-layer RNNs by an output layer whose size is the number of classes $C$. Besides, we assume that the label of $\hat{\mathbf{x}}_{i}$ equals to the label of its corresponding pixel $\mathbf{x}$. Then, we will have $N\times k$ samples. These samples are used to train convolutional layers. After that, the weights of these convolutional layers are fixed and the $N$ training samples are used again to train the two-layer RNNs. Finally, the whole network is fine-tuned based on the learned parameters.

\section{Experiments}
\subsection{Data Description}
Our experiments are conducted on two HSIs, which are widely used to evaluate classification algorithms.

\begin{figure}
  \centering
  \includegraphics[scale = 0.6]{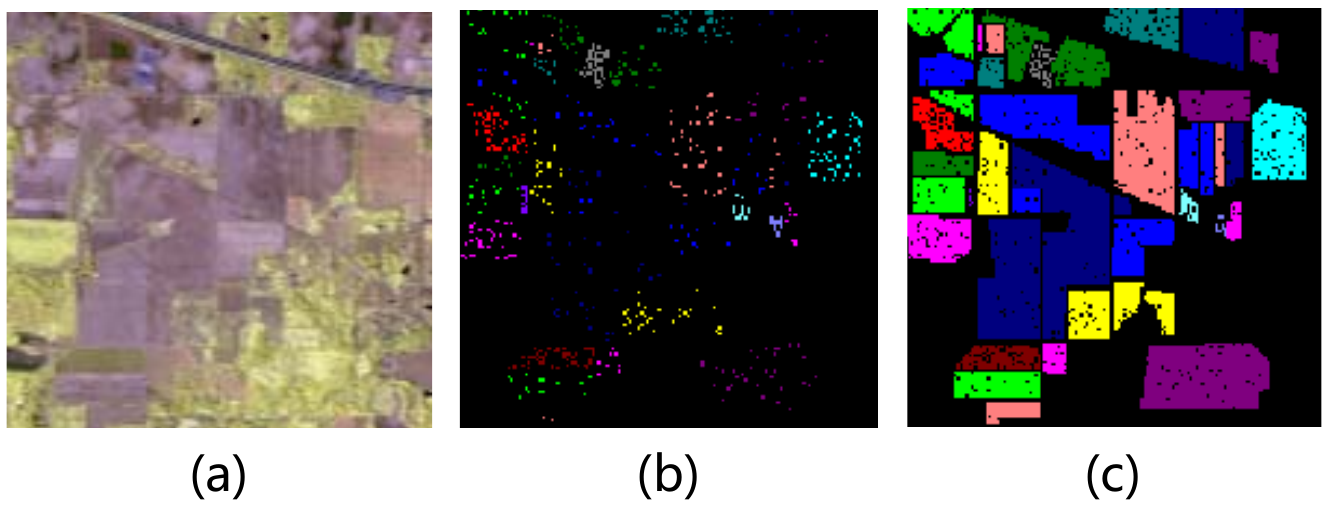}\\
  \caption{Visualization of the Indian Pines data. (a) False-color image. (b) Training data map. (c) Test data map.}\label{RGBIP}
\end{figure}

\textit{Indian Pines Data}: The first data set was acquired by the AVIRIS sensor over the Indian Pine test site in northwestern Indiana, USA, on June 12, 1992. The original data set contains 224 spectral bands. We utilize 200 of them after removing four bands containing zero values and 20 noisy bands affected by water absorption. The spatial size of the image is $145\times145$ pixels, and the spatial resolution is 20 m. The number of training and test pixels are reported in Table$~$\ref{IPData}. Fig.$~$\ref{RGBIP} shows the false-color image, as well as training and test maps of this data set.
\begin{table}
  \centering
  \caption{Numbers of training and test pixels used in the Indian Pines data set.}\label{IPData}
  \scalebox{0.9}{
  \begin{tabular}{cccc}
  \hline
     Class No. & Class Name & Training & Test\\
     \hline
     \hline
     1 & Corn-notill & 50 & 1384 \\
     2 & Corn-mintill & 50 & 784 \\
     3 & Corn & 50 & 184 \\
     4 & Grass-pasture & 50 & 447 \\
     5 & Grass-trees & 50 & 697 \\
     6 & Hay-windrowed & 50 & 439 \\
     7 & Soybean-notill & 50 & 918 \\
     8 & Soybean-mintill & 50 & 2418 \\
     9 & Soybean-clean & 50 & 564 \\
     10 & Wheat & 50 & 162 \\
     11 & Woods & 50 & 1244 \\
     12 & Building-grass-trees & 50 & 330 \\
     13 & Stone-steel-towers & 50 & 45 \\
     14 & Alfalfa & 15 & 39 \\
     15 & Grass-pasture-mowed & 15 & 11 \\
     16 & Oats & 15 & 5 \\
     \hline
     \hline
      -  & Total & 695 & 9671 \\
     \hline
   \end{tabular}
   }
\end{table}

\begin{figure}
  \centering
  \includegraphics[scale = 0.5]{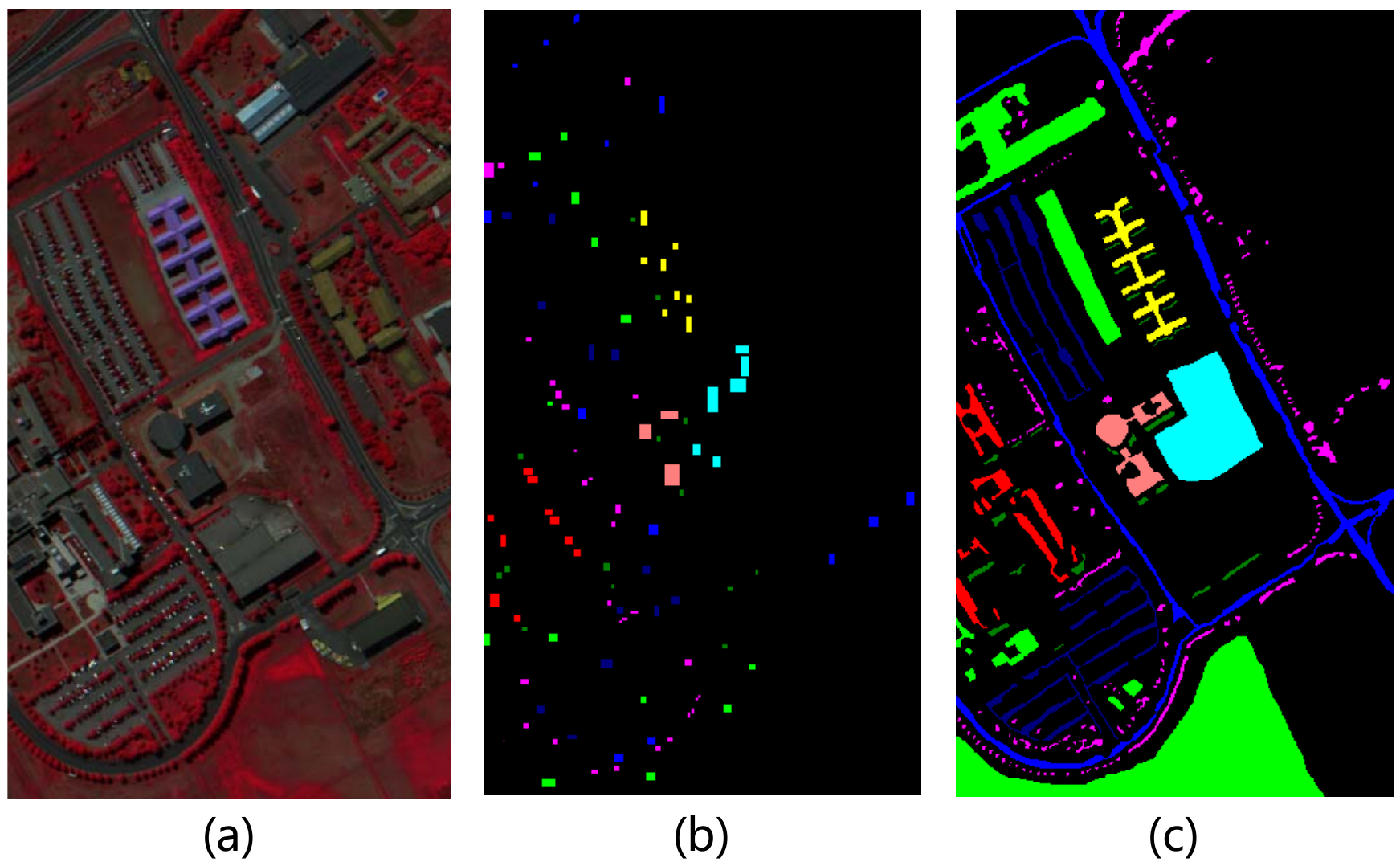}\\
  \caption{Visualization of the Pavia University data. (a) False-color image. (b) Training data map. (c) Test data map.}\label{RGBPUS}
\end{figure}

\textit{Pavia University Scene Data}: The second data set was acquired by the ROSIS sensor during a flight campaign over Pavia, northern Italy, on July 8, 2002. The original image was recorded with 115 spectral channels ranging from 0.43 $\mu m$  to 0.86 $\mu m$. After removing noisy bands, 103 bands are used. The image size is $610\times340$ pixels with a spatial resolution of 1.3 m. There are nine classes of land covers with more than 1000 labeled pixels for each class. The number of pixels for training and test are listed in Table$~$\ref{PUSData}. Their corresponding distribution maps are demonstrated in Fig.$~$\ref{RGBPUS}.

\begin{table}
  \centering
  \caption{Numbers of training and test pixels used in the Pavia University data set.}\label{PUSData}
  \scalebox{0.9}{
  \begin{tabular}{cccc}
  \hline
     Class No. & Class Name & Training & Test\\
     \hline
     \hline
     1 & Asphalt & 548 & 6631 \\
     2 & Meadows & 540 & 18649 \\
     3 & Gravel & 392 & 2099 \\
     4 & Trees & 524 & 3064 \\
     5 & Metal sheets& 265 & 1345 \\
     6 & Bare Soil & 532 & 5029 \\
     7 & Bitumen & 375 & 1330 \\
     8 & Bricks & 514 & 3682 \\
     9 & Shadows & 231 & 947 \\
     \hline
     \hline
       - & Total & 3921 & 42776 \\
     \hline
   \end{tabular}
   }
\end{table}

\subsection{Experimental Setup}
In order to highlight the effectiveness of our proposed models, we compare them with SVM, one-dimensional CNN (1D-CNN), two-dimensional CNN (2D-CNN), and the original RNN using GRU (RNN). For simplicity, the cascaded RNN model using GRUs is abbreviated as CasRNN; the two improvement methods of CasRNN based on feature-level and output-level connections are abbreviated as CasRNN-F and CasRNN-O, respectively; the spectral-spatial CasRNN is abbreviated as SSCasRNN. Some of their explanations are summarized as follows.

\begin{enumerate}
  \item SVM: The input of SVM is the original spectrum signature. We choose Gaussian kernel as its kernel function. The penalty parameter and the spread of the Gaussian kernel
      are selected from a candidate set $\{10^{-3}, 10^{-2}, \cdots, 10^{3}\}$ using a fivefold cross-validation method.
  \item 1D-CNN: The structure of 1D-CNN is the same as that in \cite{hu2015}. It contains an input layer, a convolutional layer with 20 kernels whose size is $11\times1$, a max-pooling layer whose kernel size is $3\times1$, a fully-connected layer with 100 hidden nodes, and an output layer.
  \item 2D-CNN: The structure of 2D-CNN is the same as that in \cite{chen2016}, which consists of three convolutional layers and two max-pooling layers. Please refer to Table IX in \cite{chen2016} for the design details of it.
  \item RNN: GRU is used as the basic unit of RNN. The number of hidden nodes is chosen from a candidate set $\{2^{4}, 2^{5}, \cdots, 2^{10}\}$ via a fivefold cross-validation method.
\end{enumerate}

The deep learning models are constructed with a PyTorch framework. To optimize them, we use a mini-batch stochastic gradient descent algorithm. The batch size, the learning rate and the number of training epochs are set to 64, 0.001 and 300, respectively. For SVM, we use a libsvm package in a MATLAB framework. All of the experiments are implemented on a personal computer with an Intel core i7-4790, 3.60GHz processor, 32GB RAM, and a GTX TITAN X graphic card.

The classification performance of each model is evaluated by the overall accuracy (OA), the average accuracy (AA), the per-class accuracy, and the Kappa coefficient. OA defines the ratio between the number of correctly classified pixels to the total number of pixels in the test set, AA refers to the average of accuracies in all classes, and Kappa is the percentage of agreement corrected by the number of agreements that would be expected purely by chance.

\subsection{Parameter Analysis}
\begin{figure}
  \centering
  \includegraphics[scale=0.4]{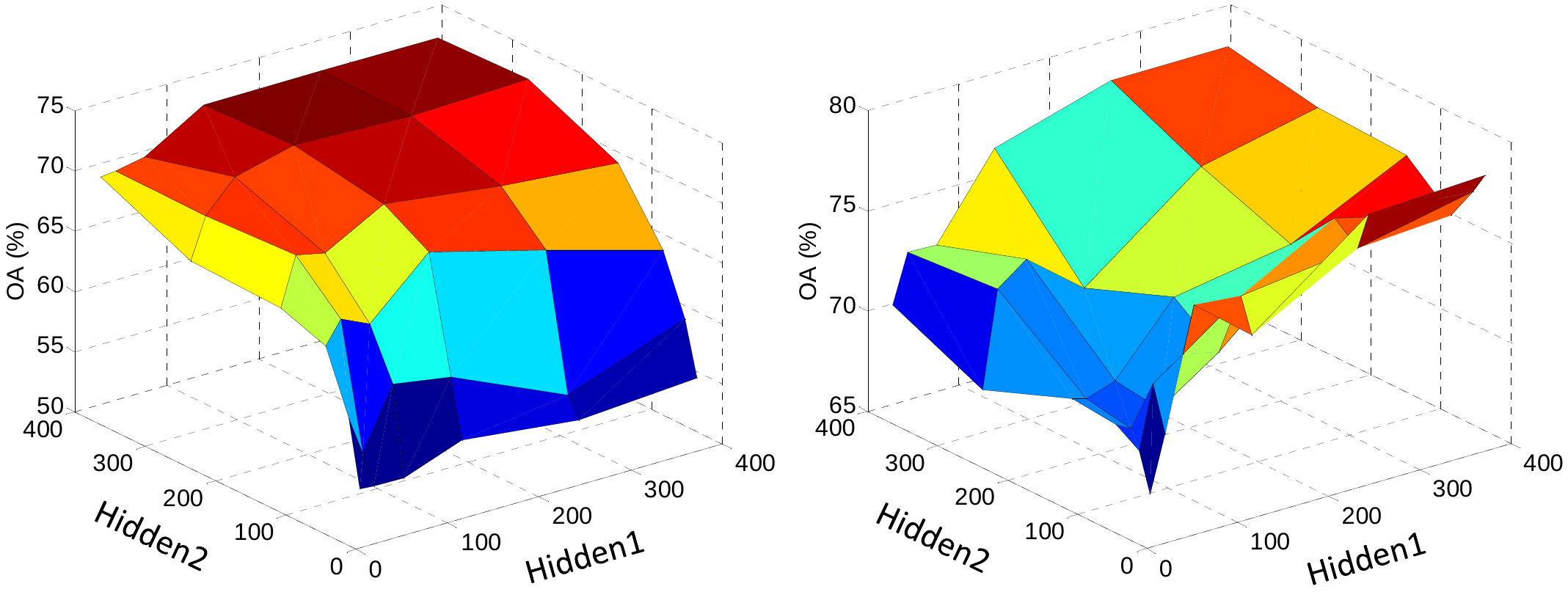}\\
  \caption{Performance of the CasRNN model with different sizes of hidden layers on the Indian Pines data (Left) and the Pavia University data (Right).}\label{CasRNNHiddenNum}
\end{figure}

\begin{figure}
  \centering
  \includegraphics[scale=0.4]{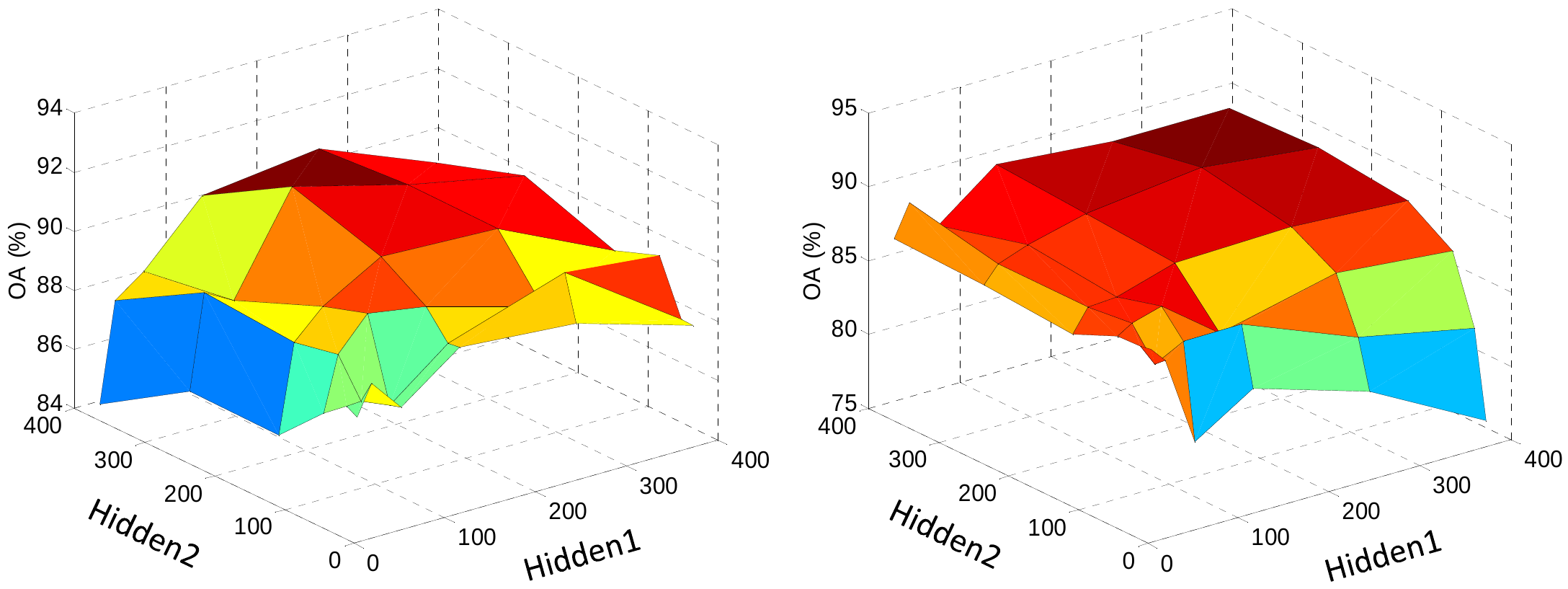}\\
  \caption{Performance of the SSCasRNN model with different sizes of hidden layers on the Indian Pines data (Left) and the Pavia University data (Right).}\label{SSCasRNNHiddenNum}
\end{figure}

\begin{figure}
  \centering
  \includegraphics[scale=0.5]{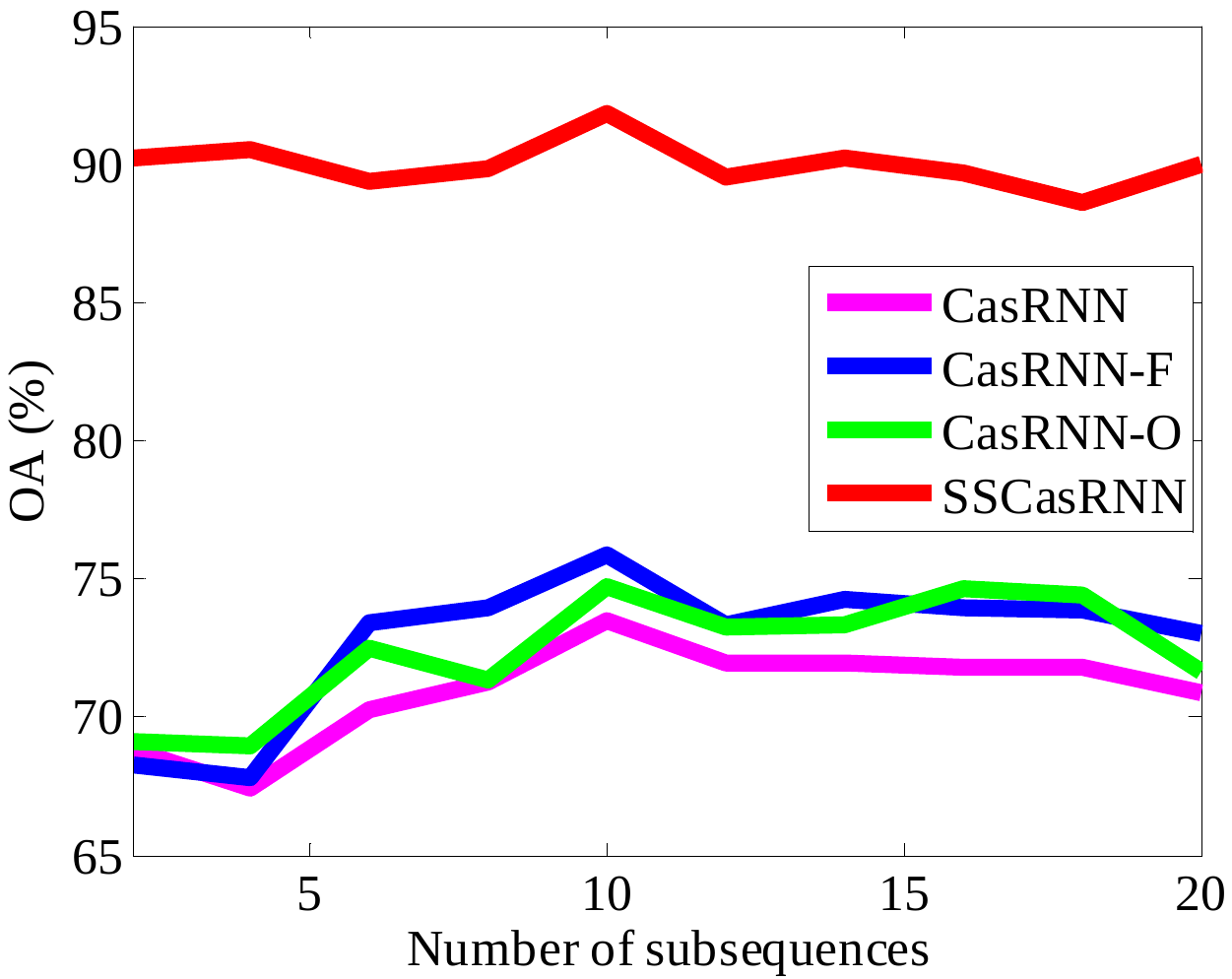}\\
  \caption{Performance of different models on the Indian Pines data with different sub-sequence numbers $l$. }\label{SubsequenceNumIP}
\end{figure}

\begin{figure}
  \centering
  \includegraphics[scale=0.5]{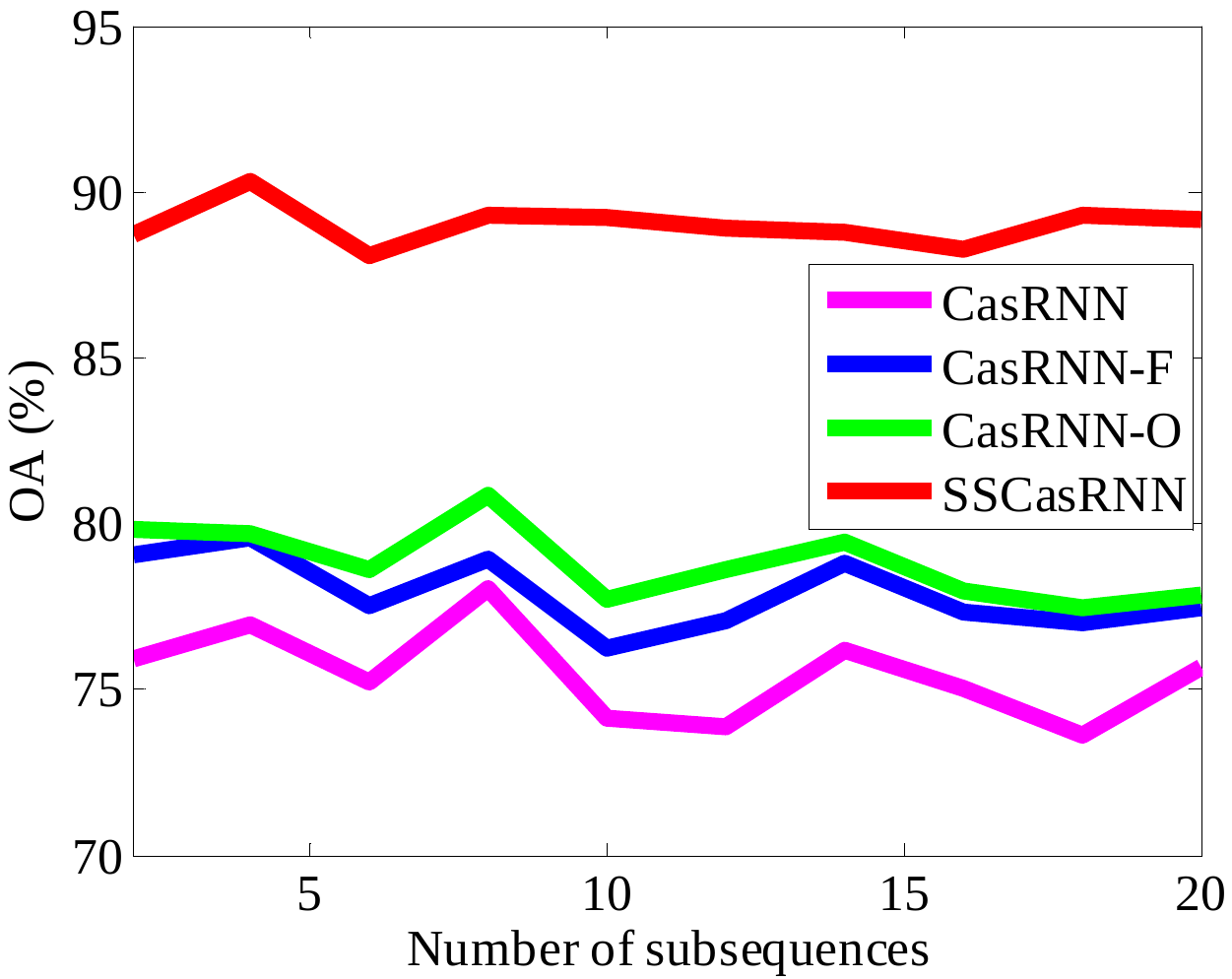}\\
  \caption{Performance of different models on the Pavia University data with different sub-sequence numbers $l$. }\label{SubsequenceNumPUS}
\end{figure}
There exist three important hyperparameters in the proposed models. They are sub-sequence numbers $l$, as well as the size of hidden layers in the first-layer RNN and the second-layer RNN. To test the effects of them on the classification performance, we firstly fix $l$ and select the size of hidden layers from a candidate set $\{16,32,64,128,256,384\}$. Then, we fix the size of hidden layers and choose $l$ from another set $\{2,4,6,\cdots,16,18,20\}$. Since the same hyperparameter values are used for CasRNN and its two improvements (i.e., CasRNN-F and CasRNN-O), we only demonstrate the performance of CasRNN here, shown in Fig.$~$\ref{CasRNNHiddenNum}. In this three-dimensional diagram, the first two axes (named $Hidden1$ and $Hidden2$) respectively correspond to the number of hidden nodes in the first-layer RNN and the second-layer RNN, while the third axis represents the classification accuracy OA. From this figure, we can observe that when $Hidden1\geq32$ and $Hidden2\geq128$, CasRNN can achieve better OA than the other values on the Indian Pines data. The best OA appears when $Hidden1=128$ and $Hidden2=256$. For the Pavia University data, OA changes a little larger than the Indian Pines data, but we can still find the best value when $Hidden1=256$ and $Hidden2=16$. Similarly, Fig.$~$\ref{SSCasRNNHiddenNum} shows OA values achieved by SSCasRNN using different hidden sizes. We can see the optimal parameter values are $Hidden1=128, Hidden2=256$ for the Indian Pines data, and $Hidden1=256, Hidden2=256$ for the Pavia University data, respectively.

\begin{table*}
  \centering
  \caption{Classification results (\%) of different models on the Indian Pines data.}\label{IPResults}
  \scalebox{0.9}{
  \begin{tabular}{ccccccccc}
    \hline
    Class No. & SVM & 1D-CNN & RNN & CasRNN & CasRNN-F & CasRNN-O & 2D-CNN & SSCasRNN \\
    \hline
    \hline
    1 & 64.31 & 61.34 & 64.74 & 68.35 & 68.93 & 68.21 & 82.51 & \textbf{86.99} \\
    2 & 70.92 & 60.33&	61.35&	64.8&67.6&67.35&88.14&\textbf{98.72}	\\
    3 & 84.78&80.43&74.46&77.17&83.7&85.87&\textbf{100}&\textbf{100}\\
    4 & 91.05 & 89.04 & 83.45 & 91.50 & 90.60 & 89.93 & \textbf{94.85} & 94.41 \\
    5 & 85.94 & 90.53 & 77.04 & 79.34 & 80.49 & 80.92 & 85.80 & \textbf{97.42} \\
    6 & 93.62 & 96.13 & 87.70 & 92.03 & 92.94 & 92.94 & 99.77 & \textbf{100} \\
    7 & 69.17 & 72.11 & 76.03 & 74.84 & 78.54 & 79.30 & 82.35 & \textbf{87.15} \\
    8 & 52.90 & 54.47 & 60.79 & 67.41 & 67.49 & 66.91 & 73.86 & \textbf{85.98} \\
    9 & 76.60 & 75.71 & 61.17 & 65.60 & 67.02 & 65.43 & 86.00 & \textbf{87.23} \\
    10 & 97.53 & 99.83 & 93.21 & 95.06 & 96.91 & 98.15 & \textbf{100} & \textbf{100} \\
    11 & 77.49 & 80.87 & 81.67 & 83.28 & 90.03 & 86.09 & 94.53 & \textbf{97.51} \\
    12 & 73.33 & 78.48 & 55.45 & 54.85 & 67.88 & 54.55 & 97.27 & \textbf{99.70} \\
    13 & \textbf{100} & 91.11 & 86.67 & 93.33 & 95.56 & 93.33 & \textbf{100} & \textbf{100} \\
    14 & 87.18 & 94.87 & 69.23 & 76.92 & 84.61 & 76.92 & 97.44 & \textbf{100} \\
    15 & 90.91 & 90.91 & 90.91 & 90.91 & 90.91 & 90.91 & \textbf{100} & \textbf{100} \\
    16 & \textbf{100} & \textbf{100} & 80.00 & \textbf{100} & \textbf{100} & 80 & \textbf{100} & \textbf{100} \\
    \hline
    \hline
    OA & 70.55 & 70.79 & 69.82 & 73.49 & 75.85 & 74.60 & 85.43 & \textbf{91.79} \\
    AA & 82.23 & 82.23 & 75.24 & 79.71 & 82.70 & 79.80 & 92.66 & \textbf{95.94} \\
    Kappa & 66.90 & 67.07 & 65.87 & 69.91 & 72.57 & 71.19 & 83.49 & \textbf{90.62} \\
    \hline
  \end{tabular}
  }
\end{table*}

\begin{figure*}
  \centering
  \includegraphics[scale = 0.85]{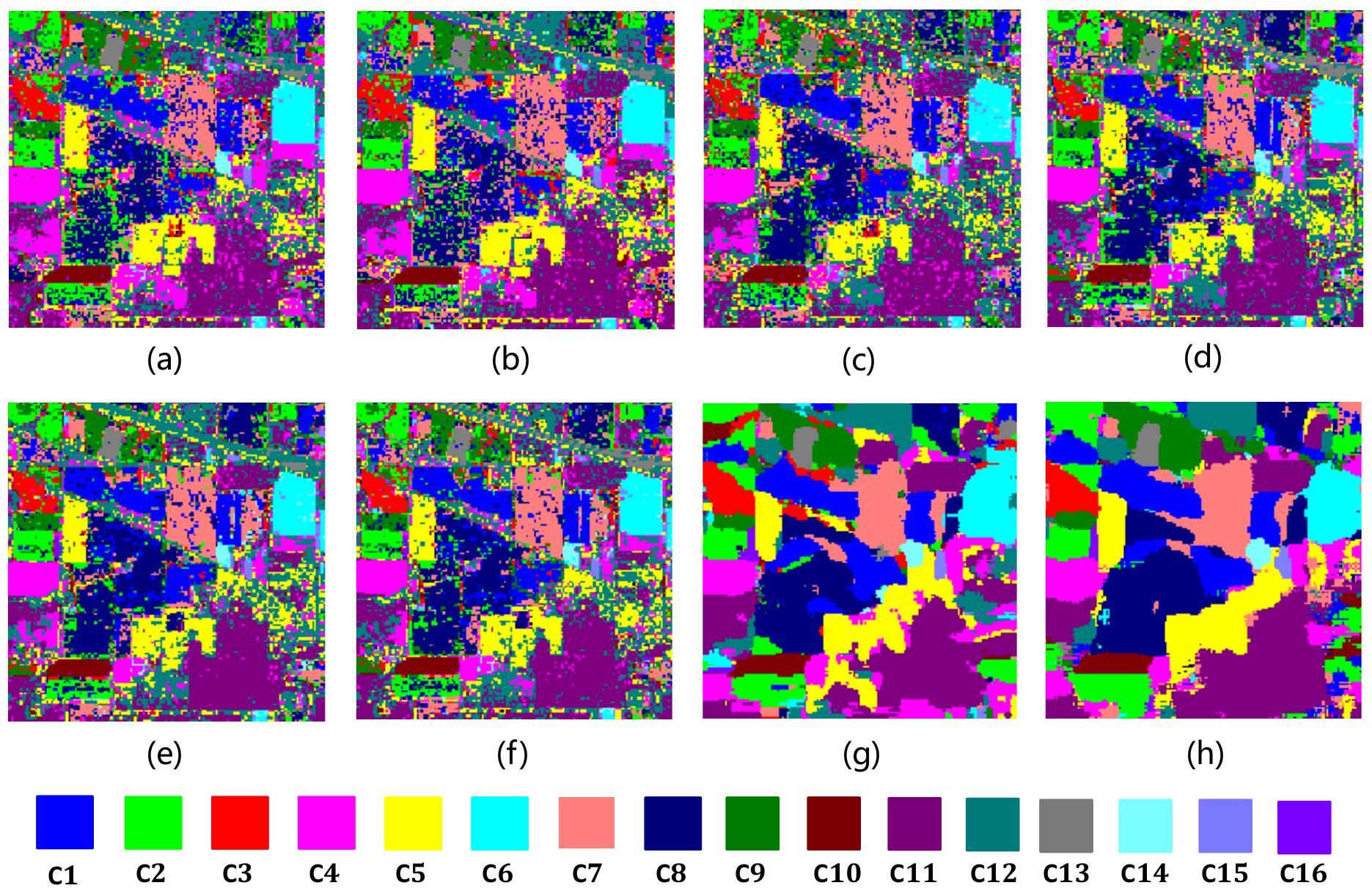}\\
  \caption{Classification maps of the Indian Pines data using different models. (a) SVM. (b) 1D-CNN. (c) RNN. (d) CasRNN. (e) CasRNN-F. (f) CasRNN-O. (g) 2D-CNN. (h) SSCasRNN. }\label{MapsIP}
\end{figure*}

\begin{table*}
  \centering
  \caption{Classification results (\%) of different models on the Pavia University data.}\label{PUSResults}
  \scalebox{0.9}{
  \begin{tabular}{ccccccccc}
    \hline
    Class No. & SVM & 1D-CNN & RNN & CasRNN & CasRNN-F & CasRNN-O & 2D-CNN & SSCasRNN \\
    \hline
    \hline
    1 & 84.74 & 80.94 & 81.51 & 82.34 & 83.56 & 83.52 & 77.39 & \textbf{89.82} \\
    2 & 64.50 & 70.37 & 62.58 & 67.13 & 70.65 & 71.37 & \textbf{98.89} & 96.06 \\
    3 & 72.56 & 77.32 & 64.65 & 60.51 & 68.75 & 64.51 & 56.74 & \textbf{78.89} \\
    4 & 97.13 & 85.93 & \textbf{98.89} & 98.63 & 98.11 & 98.43 & 92.75 & 95.89 \\
    5 & 99.55 & 99.70 & 99.26 & 99.41 & 99.55 & 99.33 & 99.78 & \textbf{100} \\
    6 & \textbf{93.30} & 93.26 & 88.90 & 84.97 & 88.29 & 89.08 & 47.27 & 57.67 \\
    7 & 91.28 & \textbf{95.41} & 92.63 & 90.60 & 76.54 & 91.13 & 80.08 & 80.53 \\
    8 & 91.99 & 84.47 & 91.04 & 92.23 & 86.04 & 93.54 & \textbf{96.69} & 96.80 \\
    9 & 95.56 & 92.08 & 95.35 & 94.40 & 95.35 & 94.72 & \textbf{96.30} & 95.99 \\
    \hline
    \hline
    OA & 78.75 & 79.55 & 76.58 & 78.03 & 79.56 & 80.86 & 86.18 & \textbf{90.30} \\
    AA & 87.85 & 86.61 & 86.09 & 85.58 & 85.21 & 87.29 & 82.88 & \textbf{87.97} \\
    Kappa & 73.62 & 74.28 & 71.02 & 72.55 & 74.31 & 75.93 & 81.22 & \textbf{86.26} \\
    \hline
  \end{tabular}
  }
\end{table*}

\begin{figure*}
  \centering
  \includegraphics[scale = 0.85]{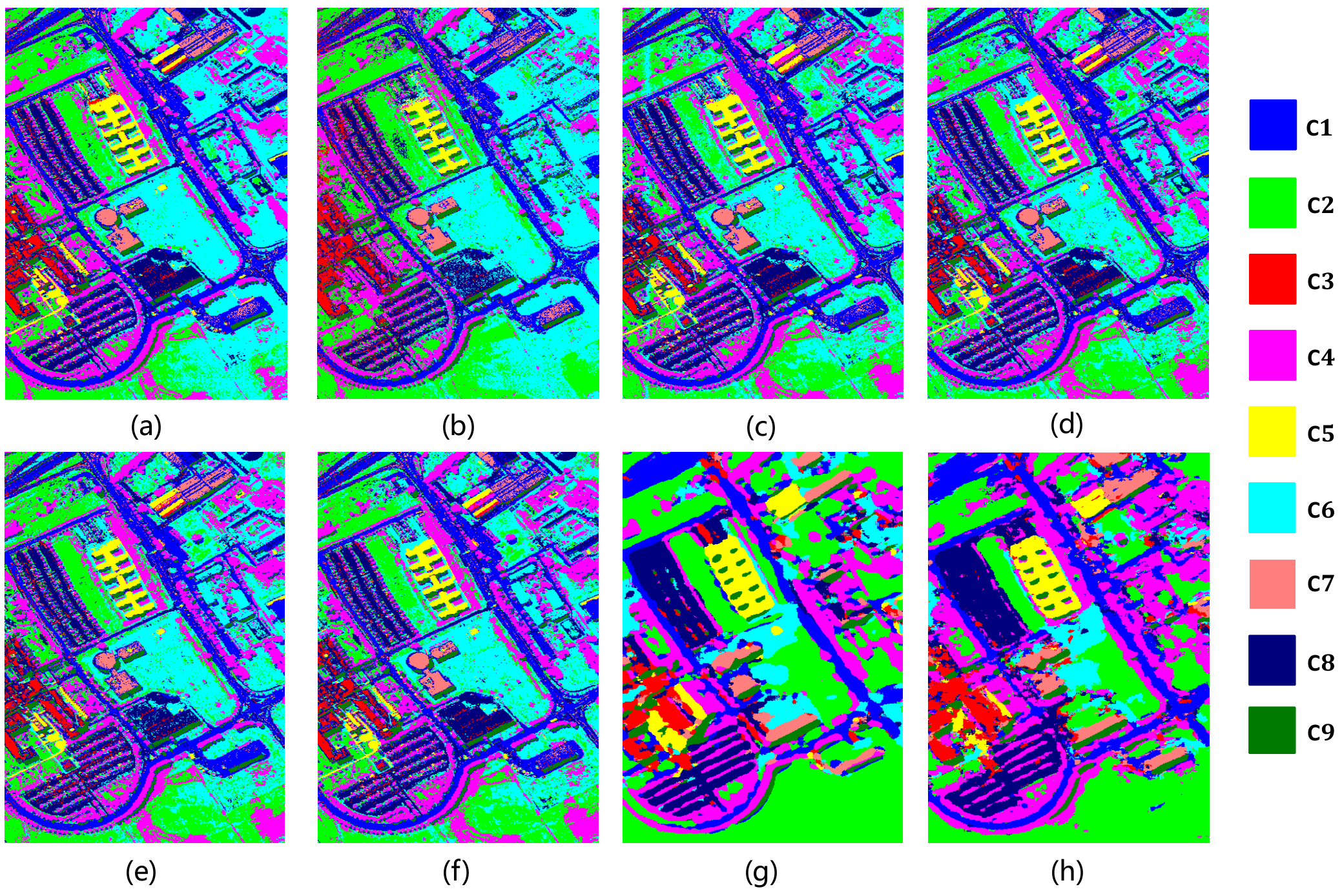}\\
  \caption{Classification maps of the Pavia University data using different models. (a) SVM. (b) 1D-CNN. (c) RNN. (d) CasRNN. (e) CasRNN-F. (f) CasRNN-O. (g) 2D-CNN. (h) SSCasRNN. }\label{MapsPUS}
\end{figure*}

Fig.$~$\ref{SubsequenceNumIP} and Fig.$~$\ref{SubsequenceNumPUS} evaluate the effects of $l$ on classifying the Indian Pines and the Pavia University data sets, respectively. In these figures, different colors represent different models. They are CasRNN, CasRNN-F, CasRNN-O and SSCasRNN. As $l$ increases, OAs achieved by these models tend to increase firstly and then decrease. Given the same $l$, SSCasRNN significantly outperforms the other three models. For the Indian Pines data, the maximal OAs of four models appear at the same $l$, so their optimal $l$ values are set as 10. Different from the Indian Pines data, four models have different optimal $l$ values on the Pavia University data. As shown in Fig.$~$\ref{SubsequenceNumPUS}, the optimal $l$ value is 4 for SSCasRNN, and 8 for the other three models.

\subsection{Performance Comparison}
In this section, we will report quantitative and qualitative results of our proposed models and their comparisons with the other state-of-the-art models. Table$~$\ref{IPResults} reports the detailed classification results of different models on the Indian Pines data, including OA, AA, Kappa and class specific accuracy. The bold fonts in each row denote the best results. Several conclusions can be observed from this table. First, if we directly input the whole spectral bands into RNN, its OA, AA and Kappa values are 69.82\%, 75.42\% and 65.87\%, respectively, which are all lower than those achieved by SVM and 1D-CNN models. This indicates that RNN cannot fully explore the long-term spectral sequence of HSIs. On the contrary, considering the redundant and complementary properties of spectral signature, our proposed model CasRNN can improve the performance of RNN by 4 percents, thus outperforming SVM and 1D-CNN. Second, compared to CasRNN, CasRNN-F and CasRNN-O can obtain better results, which validates the effectiveness of the two improvement strategies. In terms of each class accuracy, CasRNN-F almost increases all of them in comparison with CasRNN, so it might be more powerful than CasRNN-O on the Indian Pines data. Third, compared to spectral classification models, 2D-CNN significantly improves the classification results by about 10 percents. It means that the consideration of spatial information is very important on the Indian Pines data, because there are many large and homogeneous objects shown in Fig.$~$\ref{RGBIP}(c). By incorporating the spatial information into CasRNN model, our proposed model SSCasRNN can further increase the performance to above 90 percents. Besides, it can obtain highest accuracies in 15 different classes, which sufficiently certifies the effectiveness of SSCasRNN.

In addition to the quantitative results, we also visualize classification results of different models shown in Fig.$~$\ref{MapsIP}. Different colors in this figure correspond to different classes. Compared to the groundtruth map in Fig.$~$\ref{RGBIP}(c), spectral classification models (i.e., SVM, 1D-CNN, RNN, CasRNN, CasRNN-F and CasRNN-O) have many outliers in the classification map due to the spectral variability of materials. This phenomenon can be alleviated by 2D-CNN, because it makes use of the spatial contextual information instead of the spectral information. For homogeneous regions, especially large objects, 2D-CNN performs very well. However, it will easily result in an over-smoothing problem especially for small objects, as demonstrated in Fig.$~$\ref{MapsIP}(g). Different from 2D-CNN and spectral models, SSCasRNN takes advantage of spectral and spatial information simultaneously. As shown in Fig.$~$\ref{MapsIP} (h), it has significantly fewer outliers than spectral models, and retains more boundary details of objects than 2D-CNN.

Table$~$\ref{PUSResults} and Fig.$~$\ref{MapsPUS} are the classification results of different models on the Pavia University data. Similar conclusions can be observed from them. For spectral models, CasRNN is better than RNN, while CasRNN-F and CasRNN-O are superior to CasRNN. All of these models have the ``salt and pepper'' phenomenon in their classification maps. Compared to the best spectral model, 2D-CNN can improve OA and Kappa by more than 5 percents. In addition, it generates fewer outliers and leads to a more homogeneous classification map. Nevertheless, without using the spectral information, its performance is not very high, and the classification map is easily to be over-smoothed. Combining the spectral and spatial information together, our proposed model SSCasRNN can alleviate these issues. It improves OA from 86.18\% to 90.30\%, and generates more details in the classification map. However, in comparison with the Indian Pines data, the classification results achieved by SSCasRNN are still not very high. One possible reason is that there exist many small objects in the Pavia University data, which increases the difficulty in exploring spatial features.

\section{Conclusions}
In this paper, we proposed a cascaded RNN model for HSI classification. Compared to the original RNN model, our proposed model can fully explore the redundant and complementary information of the high-dimensional spectral signature. Based on it, we designed two improvement strategies by constructing connections between the first-layer RNN and the output layer, thus generating more discriminative spectral features. Additionally, considering the importance of spatial information, we further extended the proposed model into its spectral-spatial version to learn spectral and spatial features simultaneously. To test the effectiveness of the proposed models, we compared them with several state-of-the-art models on two widely used HSIs. The experimental results demonstrate that the cascaded RNN model can obtain higher performance than RNN, and its modifications can further improve the performance. Besides, we also thoroughly evaluated the effects of different hyperparameters on the classification performance of the proposed models, including the hidden sizes and the number of sub-sequences. In the future, more experiments will be conducted to validate the effectiveness of our proposed models. In addition, more powerful spectral-spatial models will be explored. Since the sizes and shapes of different objects vary, using the patches or cubes with same sizes as inputs easily leads to the loss of spatial information.

\bibliography{IEEEfull,GRUs}
\bibliographystyle{IEEEbib}

\end{document}